\documentclass[10pt]{article}

\usepackage{amsfonts}
\usepackage{amsmath}
\usepackage{bm}
\usepackage{caption}
\usepackage{float}
\usepackage{geometry}
\usepackage{graphicx}
\usepackage[none]{hyphenat}
\usepackage[hidelinks]{hyperref}
\usepackage[utf8]{inputenc}
\usepackage{subcaption}
\usepackage[most]{tcolorbox}
\usepackage{tikz}
\usepackage{tikzscale}
\usepackage{titling}
\usepackage{xcolor}
\usetikzlibrary{shapes.multipart}

\title{Dynamic survival prediction in intensive care units from heterogeneous time series without the need for variable selection or pre-processing}
\author{
Jacob Deasy\thanks{jd645@cam.ac.uk}\protect\phantom{\footnotesize 1}\thanks{University of Cambridge, Computer Laboratory, William Gates Building, 15 JJ Thomson Ave, Cambridge, CB3 0FD, UK.} \ Pietro Li\`{o}\thanksmark{2} \ and Ari Ercole\thanks{University of Cambridge, Division of Anaesthesia, Addenbrooke's Hospital, Hills Road, Cambridge, CB2 0QQ, UK.}
}
\date{}

\begin{document}
\maketitle

\vspace{-15mm}
\noindent\textbf{ABSTRACT}\vspace{1mm}\\
\textbf{Background} Intensive Care Units (ICUs) treat the most critically ill patients in the hospital. Extensive monitoring and investigation generates large quantities of data which may contain numerous, time-sensitive trends and correlations which are difficult for clinicians to systematically evaluate. In principle computers do not suffer this limitation, but the sheer complexity of the data means that this approach has previously required careful variable selection and data cleaning. This discards potentially pertinent information and is a significant barrier to implementation and generalisability in a heterogeneous ecosystem of Electronic Health Records (EHRs). In this work we present a simplified machine learning pipeline and model that uses the entire uncurated EHR for prediction of in-hospital mortality at arbitrary time intervals, using all available chart, lab and output events, without the need for pre-processing or feature engineering.\\

\noindent\textbf{Methods} Data for more than 45,000 American ICU patients from the MIMIC-III database were used to develop an ICU mortality prediction model. All chart, lab and output events were treated by the model in the same manner inspired by Natural Language Processing (NLP). Patient events were discretized by percentile and mapped to learnt embeddings before being passed to a Recurrent Neural Network (RNN) to provide early prediction of in-patient mortality risk. We compared mortality predictions with the Simplified Acute Physiology Score II (SAPS II) and the Oxford Acute Severity of Illness Score (OASIS). Data were split into an independent test set (10\%) and a ten-fold cross-validation was carried out during training to avoid overfitting.\\

\noindent\textbf{Findings} 13,233 distinct variables with heterogeneous data types were included without manual selection or pre-processing. Recordings in the first few hours of a patient's stay were found to be strongly predictive of mortality, outperforming models using SAPS II and OASIS scores within just 2 hours and achieving a state of the art Area Under the Receiver Operating Characteristic (AUROC) value of 0$\cdot$80 (95\% CI 0$\cdot$79-0$\cdot$80) at 12 hours vs 0$\cdot$70 and 0$\cdot$66 for SAPS II and OASIS at 24 hours respectively. Our model achieves a very strong performance of AUROC 0$\cdot$86 (95\% CI 0$\cdot$85-0$\cdot$86) for in-patient mortality prediction after 48 hours on the MIMIC-III dataset. Predictive performance increases over the first 48 hours of the ICU stay, but suffers from diminishing returns, providing rationale for time-limited trials of critical care and suggesting that the timing of decision making can be optimised and individualised.\\

\noindent\textbf{Interpretation} Simple association of all events with a patient stay ID and a timestamp allows for the incorporation of patient-specific data. More flexible model intake leads to increased model capacity when differentiating mortality risk between patients due to the expressive power of non-physiological records. Deep learning models can outperform traditional severity scores in this domain and can increasingly be understood through both time and variable space to generate novel connects between ICU patient features. Interpretable and dynamic machine learning models are key in critical care, and our results outline how to progress towards the transformation of advanced models into time-sensitive, transparent, and dependable clinical tools.\\

\noindent\textbf{Funding} UK Medical Research Council and the Raymond and Beverley Sackler Foundation.

\pagebreak

\noindent\textbf{INTRODUCTION}\vspace{1mm}\\
\noindent Clinicians in the intensive care unit (ICU) frequently need to make outcome- and time-critical decisions. To this end, ICU patients are routinely highly investigated and monitored, providing data to alert health care providers to deterioration and optimally inform such decision making. As a result, the ICU has higher data volume, variety and velocity than any other clinical setting. Such considerations make it a challenge to fully appreciate all the information available as well as inter-temporal relationships between clinical variables, particularly in the context of complex antecedent events and disease histories in a dynamically evolving environment.\\

On a day-to-day basis, it is unlikely that clinicians fully, systematically and robustly appraise all the information routinely available to them. For example, the fact that it is possible for algorithms with less data to outperform clinicians for example when considering which ICU patients can successfully be stepped down \cite{desautels2017prediction}, suggests that not all predictive power available is exploited in decision making. This may compromise outcomes. In more extreme circumstances, human factors research has demonstrated that being overwhelmed by data leads to unconscious, and therefore potentially sub-optimal, exclusion of available information to once again make rapid decision making tractable, and this is well illustrated in the ICU by the phenomenon of `alarm fatigue' \cite{graham2010monitor,drew2014insights} which may compromise patient safety acutely.\\

Such considerations suggest that critical care is an area that is highly likely to benefit from successful exploitation of EHRs to assist clinicians in making optimally informed decisions \cite{goldstein2017opportunities,johnson2016machine}. However, any technology to help the clinician must deal with data that is highly heterogeneous both in type (ranging from continuous variables such as laboratory results to event data such as interventions, drug administrations or clinical assessments) and in sampling (which ranges from demographic parameters through to time-series data). Furthermore the data may be subject to variable or irregular sampling and possibly informative missingness \cite{che2018recurrent}. \\

A final challenge is that ICU admissions are dynamic with prognostic accuracy which changes over time \cite{meiring2018optimal}. Clinicians must continually reconsider prognosis likelihoods while attending to the readings of multiple patients, each of whom may have a vast array of differing predictor variables. Despite this, temporal changes in prognostic ability during ICU admission has received relatively little attention to-date. \\

Recently, the authors of \cite{rajkomar2018scalable} presented the first attempt to use \emph{all available} patient data by mapping the entire EHR to a highly curated set of predictor variables, structured inline with the categories of data available. Although this method achieved strong performance, risk assessment was EHR format-specific, static, and reliant on an ensemble of diverse model structures. Nevertheless, we conjecture that the embedding mechanism outlined therein could be applied to the higher resolution setting of the ICU to usefully provide a dynamic estimate of survival probability as a composite surrogate for patient state. However, we extend this method to make it time sensitive; allowing predictions to be made at arbitrary points, optimally utilising all information available at the time. Our objective was to design and implement a pipeline for prediction which incorporated \textit{all} chart, lab and output events in the same way without the need for variable selection, curation or pre-processing. Furthermore, we sought to evaluate model explainability by ranking the features the model had paid attention to when making its prediction.\\

\noindent\textbf{METHODS}\vspace{1mm}

\noindent\textbf{Data sources}

\noindent We built and validated a computational clinical support model using retrospective analysis of adult patient data using the MIMIC-III database \cite{johnson2016mimic}. The database contains high-resolution patient data, including: demographics, vital sign time-series, laboratory tests, illness severity scores, medications and procedures, fluid intake and outputs, clinician notes, and diagnostic coding.  The median age of adult patients is 65$\cdot$8 years ($Q_1-Q_3$: 52$\cdot$8–77$\cdot$8), 55$\cdot$9\% patients are male, and in-hospital mortality is 11$\cdot$5\% \cite{johnson2016mimic}. Between the two EHR systems that comprise MIMIC-III, CareVue and MetaVision, in total the overall dataset contains 330,712,483 chart events, 27,854,055 lab events and 4,349,218 output events.\\

\noindent\textbf{Procedures}\\
We follow the patient exclusion criteria outlined in the MIMIC-III benchmark \cite{harutyunyan2017multitask}, by excluding patients if the patient is less than 18 years old at the time of ICU admission or the patient's mortality is not documented, and excluding events if the event cannot be mapped to an ICU stay or the event cannot be mapped to a hospital admission. Patients with multiple stays in the ICU were included in the dataset. We did not deliberately include patient demographic information, although it is often recorded as part of the chart events just after admission. On the basis of outcome records (i.e. survival or non-survival), we calculated in-hospital mortality - defined as death on any ward during hospital admission.\\

Unlike traditional approaches, we retain all of the chart, lab and output events for each stay without any data cleaning, outlier removal or domain-specific knowledge. The processing we perform is enough to assign a patient, a stay-ID and a timestamp to each event - a process which is independent of EHR data formatting or structure. Our model takes the entire patient timeseries as input, regardless of event type, frequency or cardinality. We note that, because we do not select for clinical variables, after event association with patient stays, our EHR dataset contains 208,572,237 events instead of the 31,868,114 employed in \cite{harutyunyan2017multitask} and all subsequent papers relying on the MIMIC-III benchmark. Due to the increased number of variables used by our model, after processing we also have a higher number of patients and stays available, supporting the evidence in favour of models which can incorporate broad EHR data.\\

\begin{figure}
    \centering
    \begin{subfigure}[b]{0.25\linewidth}
        \centering
        \includegraphics[width=\textwidth]{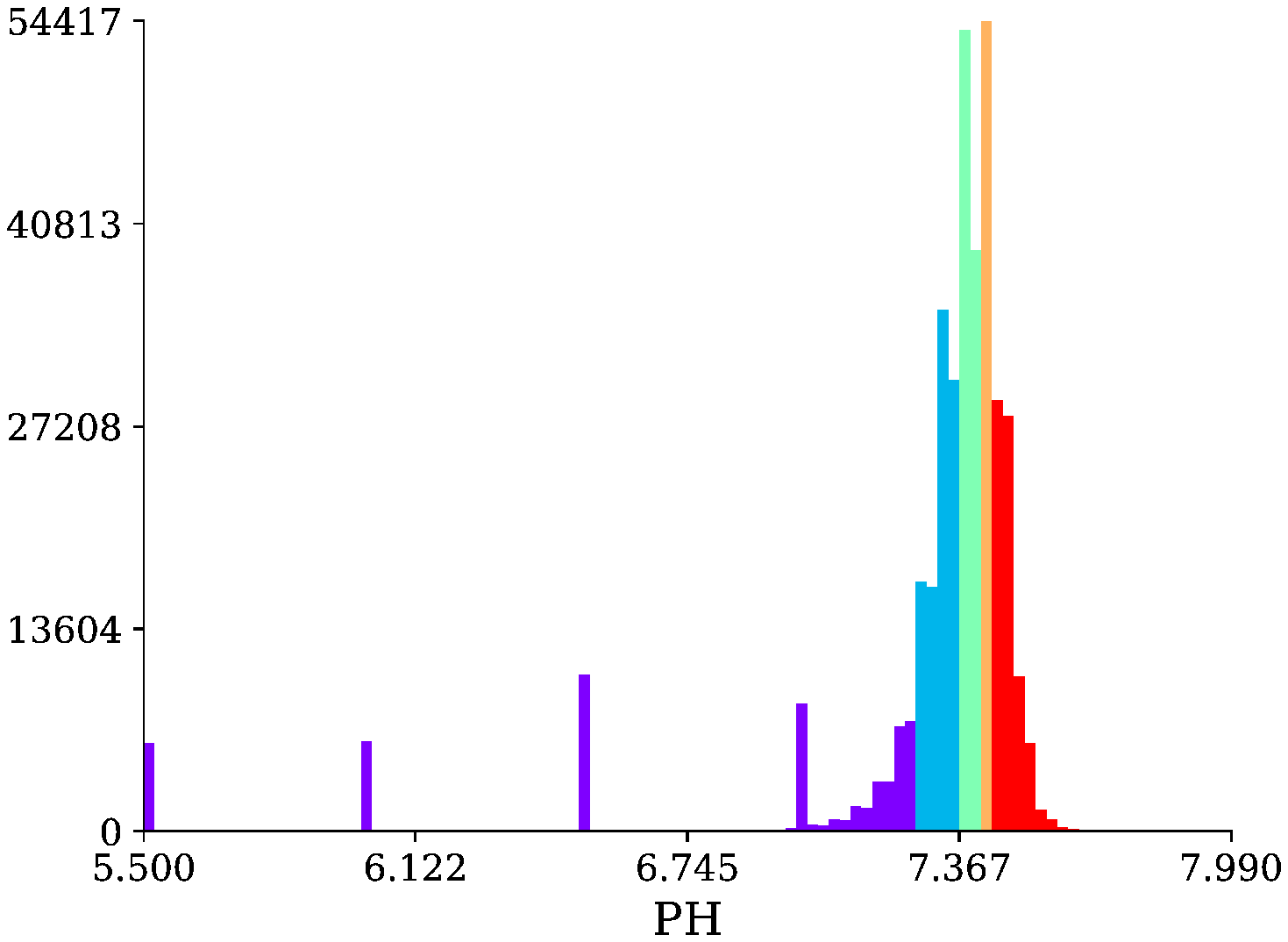}
        \caption{pH distribution, 5 bins.}
        \label{fig:ph_5bins_dist}
    \end{subfigure}%
    \begin{subfigure}[b]{0.25\linewidth}  
        \centering 
        \includegraphics[width=\textwidth]{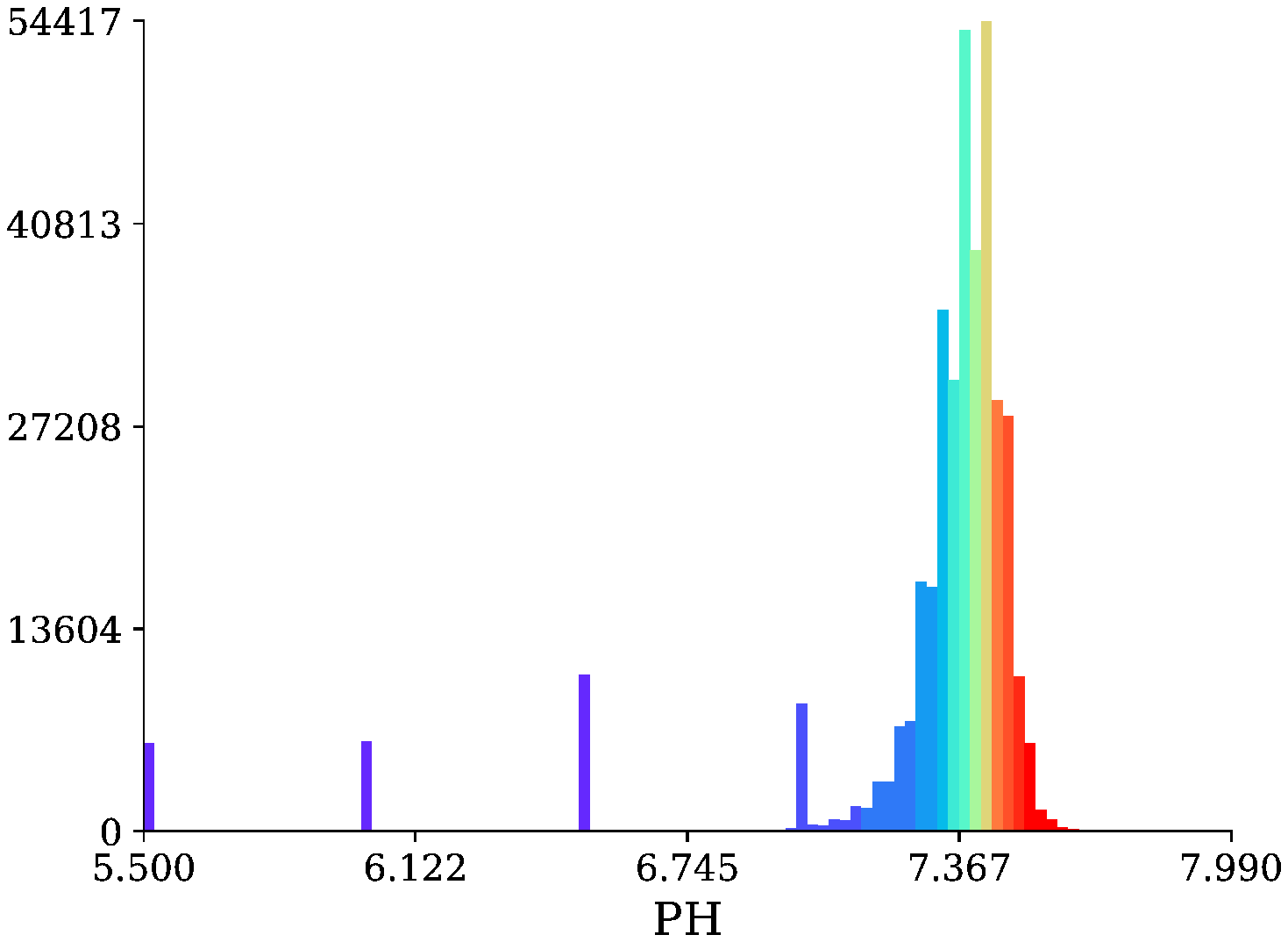}
        \caption{pH distribution, 20 bins.}
        \label{fig:ph_20bins_dist}
    \end{subfigure}%
    \begin{subfigure}[b]{0.25\linewidth}   
        \centering 
        \includegraphics[width=\textwidth]{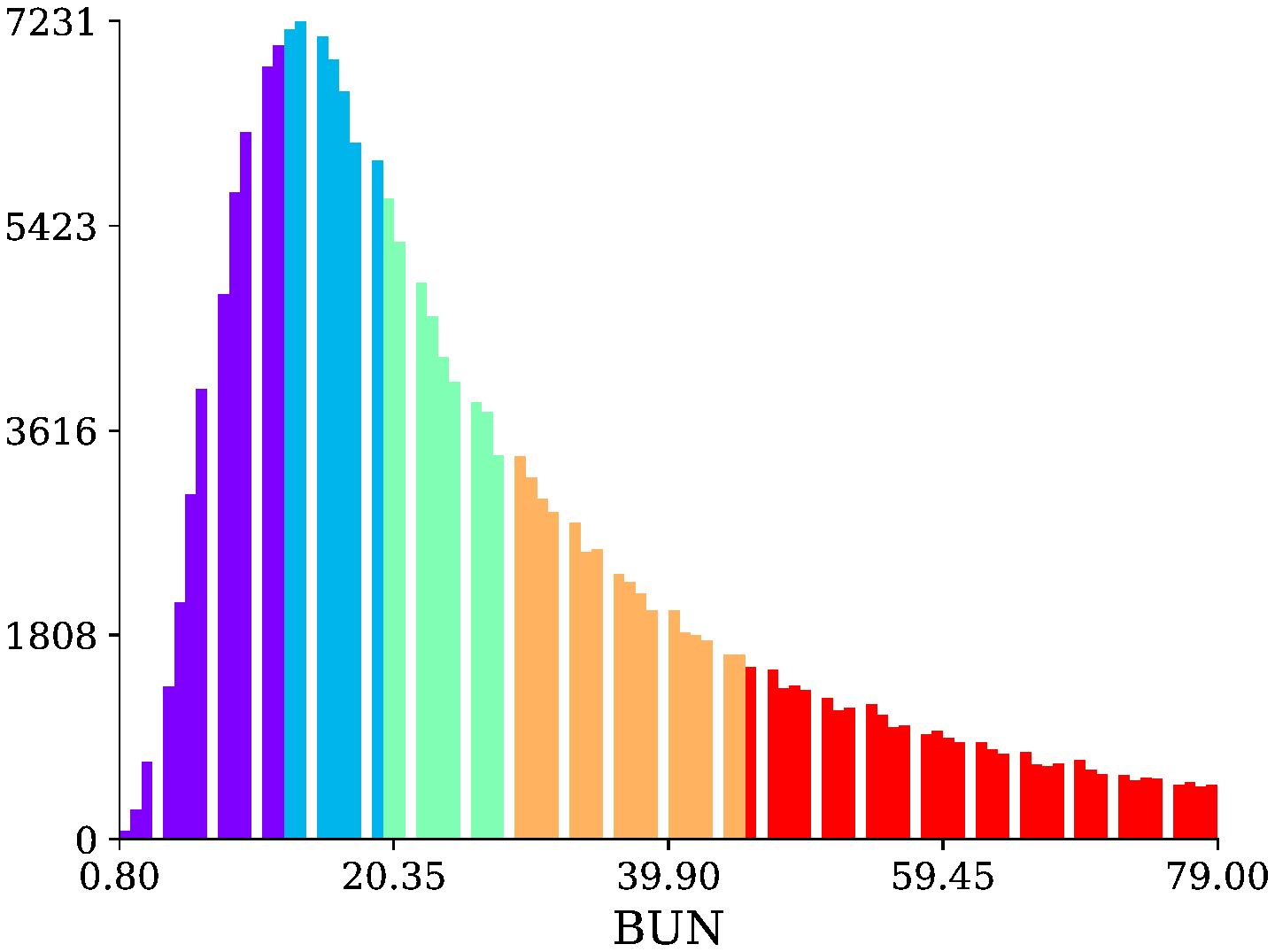}
        \caption{BUN distribution, 5 bins.}
        \label{fig:bun_5bins_dist}
    \end{subfigure}%
    \begin{subfigure}[b]{0.25\linewidth}   
        \centering 
        \includegraphics[width=\textwidth]{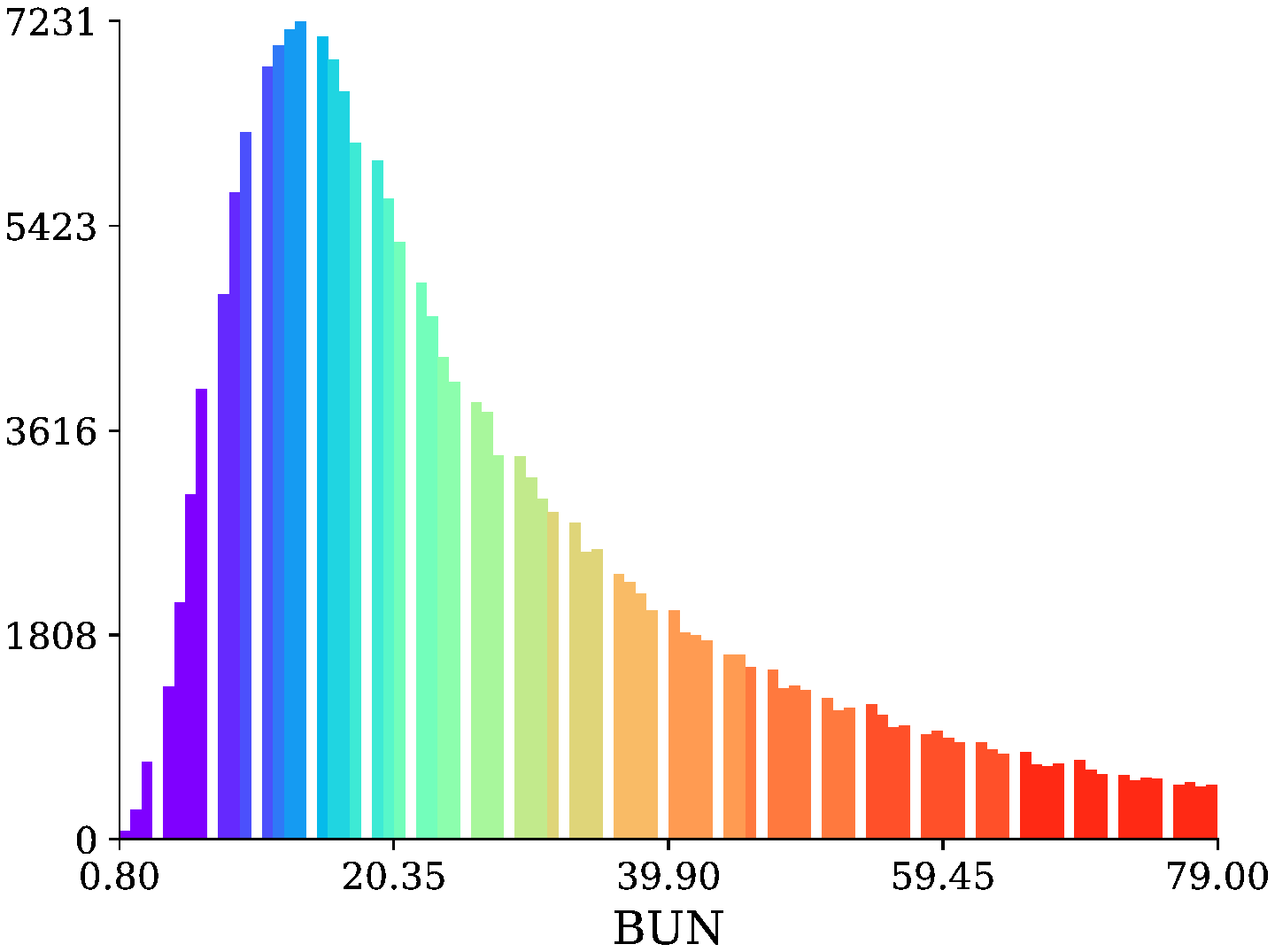}
        \caption{BUN distribution, 20 bins.}
        \label{fig:bun_20bins_dist}
    \end{subfigure}
    \caption{Representative examples of the distribution of blood pH and blood urea nitrogen (BUN) illustrating the effect of discretisation by `binning'. In our method, continuous variables are discretised by distribution frequency so that all data types can be handled in the same way in the model. Colours exemplify 5 or 20 discrete categories established by our pipeline for any continuous variable, demonstrating outlier category detection and the increased granularity in populous intervals found by percentile-based quantization. For BUN the distribution is fairly continuous and binning creates a representation which naturally encodes the concepts of `high' or `low' within the distribution. For variables such as pH however, the discretisation also places artefactual values into one (a) or more (b) `outlier' bins. If artefacts are random, the model should be able to learn that such data points have no predictive value and can therefore be ignored.}
    \label{fig:label_hists}
\end{figure}

To distinguish between discrete and continuous variables, we label events by whether their values can be converted to a floating point number. This captures all integer or decimal events, such as heart rate or blood pH, and ignores discrete labels (E.g. `Code Status Full Code'). Additionally, unusual or faulty cases, such as readings with multiple decimal places are designated as discrete, making our model robust to and aware of consistent errors which potentially correlate with patient outcomes. For missing readings, we use a zero vector embedding, but note that alternatively using a dense embedding in our model simply induces a bias towards a different point in the latent space. As a result, for both invalid and missing values, our model capitalises on information present from `informative missingness' \cite{che2018recurrent,malone2018learning}. We tokenize continuous values by quantizing them into discrete bins by percentile - our default model uses 20 such bins. Multiple quantiles reinforce the robust nature of our model, as any outliers (e.g. a regular mistake in Blood pH data is to report a pH of 5$\cdot$5, see figure~\ref{fig:ph_5bins_dist}) are likely to be contained in tokens at the periphery of a variable's distribution, and the model can learn to ignore these extremes . Examples of this tokenization and quantization procedure are given in Table~\ref{tab:labels} and Figure~\ref{fig:label_hists}.\\

\begin{table}
    \renewcommand*{\arraystretch}{1.3}
    \footnotesize
    \centering
    \begin{tabular}{|l|l|l|}
         \hline
         \textbf{LABEL} & \textbf{VALUE} & \textbf{TOKEN} \\ \hline
         Eye Opening & 4 Spontaneously & Eye Opening 4 Spontaneously \\ \hline
         Heart Rate & 84 & Heart Rate\_8 \\ \hline
         Code Status & Full Code & Code Status Full Code \\ \hline
    \end{tabular}
    \caption{Examples of discrete token creation alongside percentile-based quantization and tokenization of continuous variable. Discrete variables remain discrete, while values that can be converted to floating point numbers are considered continuous and separated into percentile-based bins.}
    \label{tab:labels}
\end{table}

\begin{figure*}
    \centering
    \input{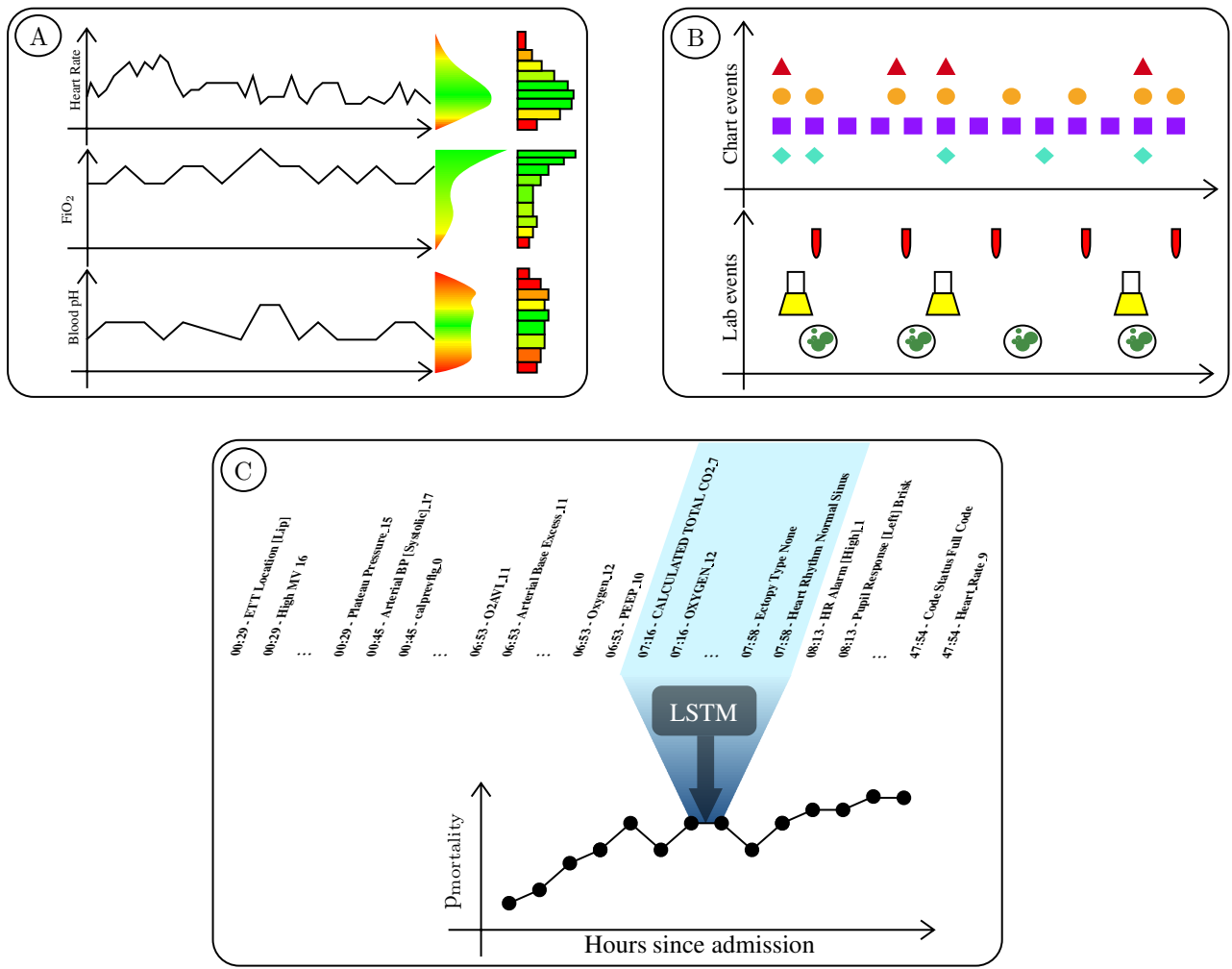}
    \caption{Diagram of our model. In panel A, continuous data, such as pH, FiO$_2$ and heart rate, are discretized into percentile-based bins, allowing both continuous features to be mapped to an embedding. In panel B, we see both discrete chart and lab events, from a diverse range of investigations, which are equivalently aggregated by the model. In panel C, both continuous and discrete events from one hour of patient timeseries data are embedded and aggregated, using a learnt variable importance ranking. The weighted average embedding is then used as input to a LSTM recurrent neural network which generates an updated dynamic prediction of in-hospital mortality probability each hour and updates an internal representation of the patient state. Dynamic prediction allows for continual patient monitoring as new data is accumulated and used to update outcome probabilities. Laboratory values, physiological readings, and admission information was from the first 48 hours after ICU admission. [LSTM: Long short-term memory recurrent neural network architecture; FiO$_2$: Fraction of inspired oxygen]}
    \label{fig:diagram}
\end{figure*}

\noindent\textbf{Model development}\\
Patient data were considered as a multivariate timeseries defined by: the times when patient events were recorded, the sequence of indexes mapped from each patient's discretized timeseries, and the set of outcomes for each patient episode. In the case of mortality prediction, patient survival was labelled with the value 1, and patient death was labelled 0. Our baseline model used snap shots that contained all chart, lab and output events for each hour of a patient's timeseries. RNN models were trained on chart, lab and output data that met the inclusion criteria within the chosen time period. A 10-fold cross-validation scheme was used to prevent overfitting, in addition to an independent test set. Data for ICU admissions were split into a training set (90\%) and an independent test set (10\%). For each cross-validation fold, training data were divided into a training set and a 1000 patient validation set, with the split stratified by survival to ensure balanced training. A LSTM RNN architecture of depth one with a single output head per time step was trained by backpropagation on the training set. As the highly variable and patient-specific nature of EHR data is prone to overfitting, training was stopped once the validation set AUROC plateaued for more than 5 epochs. During training, model performance on the validation set was continually assessed, and the optimal model for each cross-validation fold was selected from the epoch with the best validation set performance.\\

As EHR data is known to be broad and highly variable, even after all continuous variables in the relatively small MIMIC-III database were binned into 20 discrete percentile categories, the number of variables remaining was still 58,704. Therefore, traditional methods, such as learning a transition dynamics matrix or one-hot encoding all of the input variables, would be prohibitively expensive and prone to overfitting. To circumvent this potential over-parameterization, our model takes inspiration from natural language processing (NLP). Much like in NLP, where input and output vocabularies for translation are often very large, in the case of many thousands of medical tokens it is more computationally efficient to let the model learn a low-dimensional vector representation of each token.\\

We map tokens recorded in the model's medical vocabulary according to
\begin{align}
    \overbrace{
    \begin{bmatrix}
        \textrm{Eye Opening 4 Spontaneously}\\
        \textrm{Heart Rate\_8}\\
        \textrm{Code Status Full Code}\\
        \vdots
    \end{bmatrix}
    }^{\textrm{Discrete event tokens in one hour}}
    \to
    \overbrace{
    \begin{bmatrix}
        \textrm{-0$\cdot$22,\ 0$\cdot$34,\ldots,\ 0$\cdot$83} \\
        \textrm{0$\cdot$52,\ 0$\cdot$03,\ldots,\ -0$\cdot$28} \\
        \textrm{-0$\cdot$21,\ 0$\cdot$17,\ldots,\ 0$\cdot$89} \\
        \vdots
    \end{bmatrix}
    }^{\textrm{Corresponding learnt vectors}}.
\end{align}

The size of the embedding vectors was optimized via grid search over values 16, 32 and 48. Embedding dropout \cite{srivastava2014dropout,gal2016theoretically} was applied to regularize the network and prevent the model from overfitting to strongly predictive tokens which may not be available for all patients. For each patient, we allow a maximum of 10,000 events over the initial 48 hours of their stay in the ICU. For the few patients who have more than 10,000 events, we extract their final 10,000 tokens. We tested using an average vector per time period, vectors for both tokens and time period, and aggregating vectors with learnt weights. The final method performed on par with the first two, but has the added advantage of producing a ranked list of variable importance after model training - beneficial for understanding what information the model is prioritising. Therefore, we aggregate each patient timeseries snapshot according to
\begin{align}
    \overbrace{
    \begin{bmatrix}
        w_{0}, w_{1}, w_{3}, \ldots
    \end{bmatrix}
    }^{\textrm{Learnt weights}}
    \overbrace{
    \begin{bmatrix}
        \textrm{-0$\cdot$22,\ 0$\cdot$34,\ldots,\ 0$\cdot$83} \\
        \textrm{0$\cdot$52,\ 0$\cdot$03,\ldots,\ -0$\cdot$28} \\
        \textrm{-0$\cdot$21,\ 0$\cdot$17,\ldots,\ 0$\cdot$89}\\
        \vdots
    \end{bmatrix}
    }^{\textrm{Learnt vectors}}
    &=
    \overbrace{
    \begin{bmatrix}
        \textrm{0$\cdot$04,-0$\cdot$52,\ldots,-0$\cdot$72}
    \end{bmatrix}
    }^{\textrm{Aggregated hourly vector}}.
\end{align}

Finally, we use a densely connected layer with a sigmoid activation function to output $p(y_{i}|X_{i})\in[0,1]$, the probability of in-patient mortality given a patient's timeseries, and optimize the parameters of our model by minimising the binary cross entropy loss
\begin{align}
    \mathcal{L}(y, \Tilde{y}) = -\sum\limits_{i=1}^{N} \sum\limits_{t=0}^{T} \overbrace{\Tilde{y}_{it}\log y_{i}}^{\textrm{Misclassified death loss}} + \overbrace{(1-\Tilde{y}_{it})\log(1-y_{i})}^{\textrm{Misclassified survival loss}},
\end{align}
across each training batch and through time. The Adam optimizer \cite{kingma2014adam} was used for training, output activation function (sigmoid), batch size 128, and learning rate (0$\cdot$0005) were kept constant across models. The number of hidden neurons (32, 64, 128 and 256 units), as well as the probability of embedding drop-out were optimized via a grid search over the models. To establish confidence intervals, we used the bootstrap algorithm \cite{efron1994introduction} with 10,000 samples of the test set performance of the cross-validation models. A model summary is presented in Figure~\ref{fig:diagram}.\\

\noindent\textbf{RESULTS}\vspace{1mm}

\noindent Data from the MIMIC-III dataset were used for this study, including 46,476 patients with 61,532 ICU stays at the Beth Israel Deaconess Medical Center in Boston, USA. We selected valid patients and stays by initially following the selection presented in \cite{harutyunyan2017multitask}, before deviating due to the increased flexibility of our model. The percent of stays with in-hospital deaths was 13$\cdot$2\% (2,797/21,139) and the proportion of long ICU stays (greater than 7 days in the ICU) was 23$\cdot$0\% (4,868/21,139). The mean and median length of stay in the ICU was 5$\cdot$97 and 3$\cdot$72 days respectively. During the first 48 hours in the ICU, an average admission had 1,268 events, drawn from a possible 2,353 unique event names. The data was split into training and test sets which had chart, lab and output events available.\\

\begin{figure}
    \center
    \includegraphics[width=0.5\linewidth]{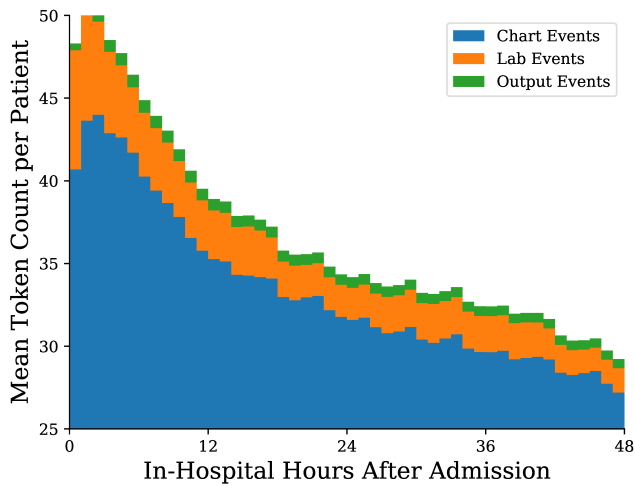}
    \caption{Distribution of mean token count per patient per hour, stratified by type, for the 48 hour cohort using 20 discrete percentile bins for continuous variables.}
    \label{fig:token_dist}
\end{figure}

Patient event types over time are summarised in Figure~\ref{fig:token_dist}. In the first few hours after admission, the average patient has between 45 and 50 readings per hour. Despite the majority of events in any given hour being chart events, in the first few hours there are also additional lab events as initial patient data is accumulated. After the initial peak in readings per hour, the number of chart events recorded declines quickly over the first 12 hours, before continuing to decline in a slower fashion for the remainder of each patient's stay. Finally, towards the 48 hour mark patient events are predominantly comprised of chart events, with an average rate of less than 30.\\

The outcomes of our discretization pipeline for two variables, blood pH (\textit{PH}) and blood urea nitrogen (\textit{BUN}), are displayed in figure~\ref{fig:label_hists}. When employing 5 percentile-based bins in figure~\ref{fig:ph_5bins_dist}, our pipeline successfully discarded several 10s of thousands of outlier readings which are the result of technical faults in recording equipment as they are not physiologically possible (e.g. blood pH below 7). Further dividing pH values into 20 percentile-based bins in figure~\ref{fig:ph_20bins_dist}, it is even possible to distinguish between different outlier categories - each of which may have slightly different correlation with patient outcomes. In the case of blood urea nitrogen, values are distributed more uniformly, with fewer outliers, so the main advantage of value discretization lies in differentiating between populous patient categories towards the distribution's centre of mass.\\

\begin{figure}
    \centering
    \begin{subfigure}[b]{0.25\linewidth}
        \centering
        \includegraphics[width=\textwidth]{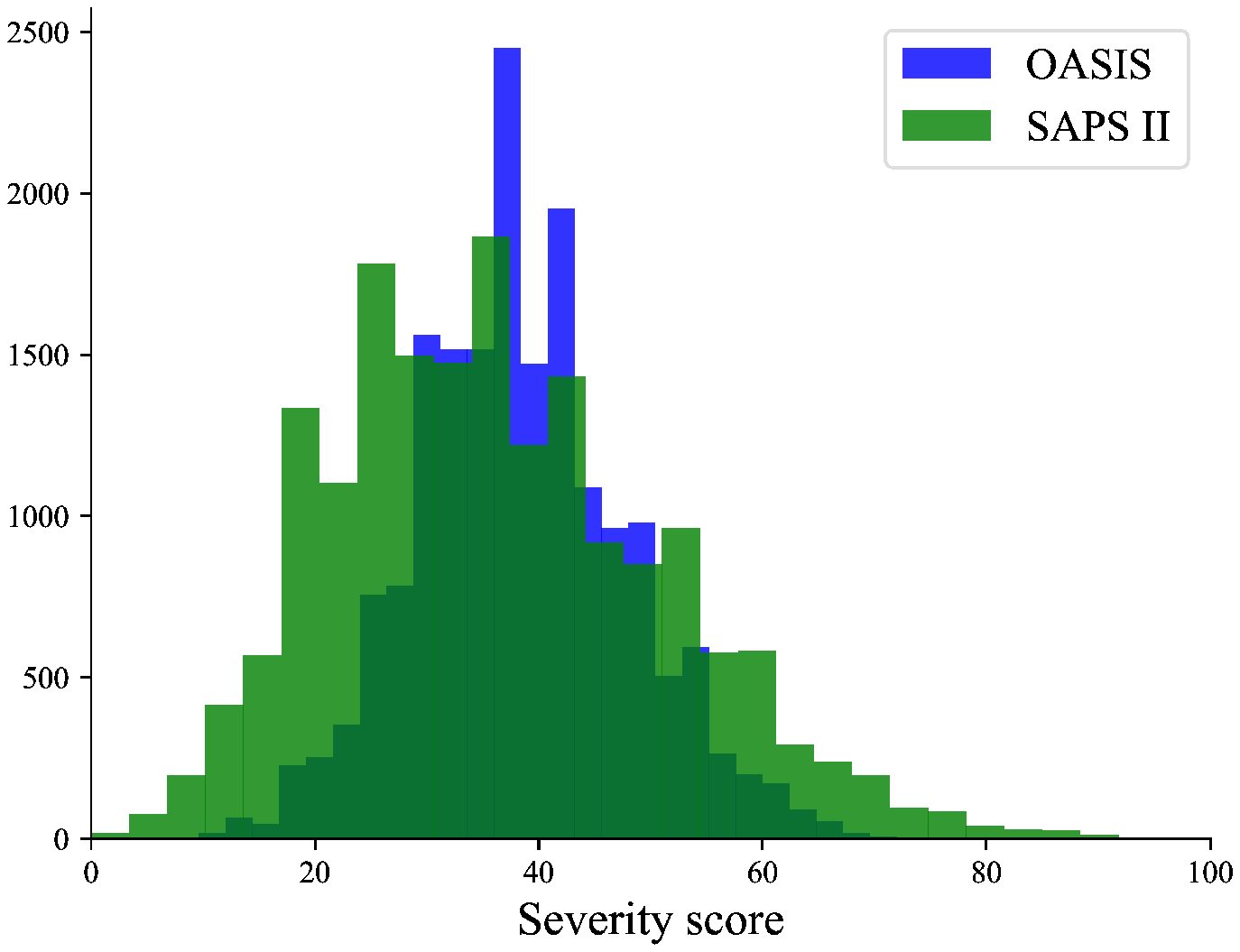}
        \caption{Train set score distribution.}
        \label{fig:train_score_dists}
    \end{subfigure}%
    \begin{subfigure}[b]{0.25\linewidth}
        \centering 
        \includegraphics[width=\textwidth]{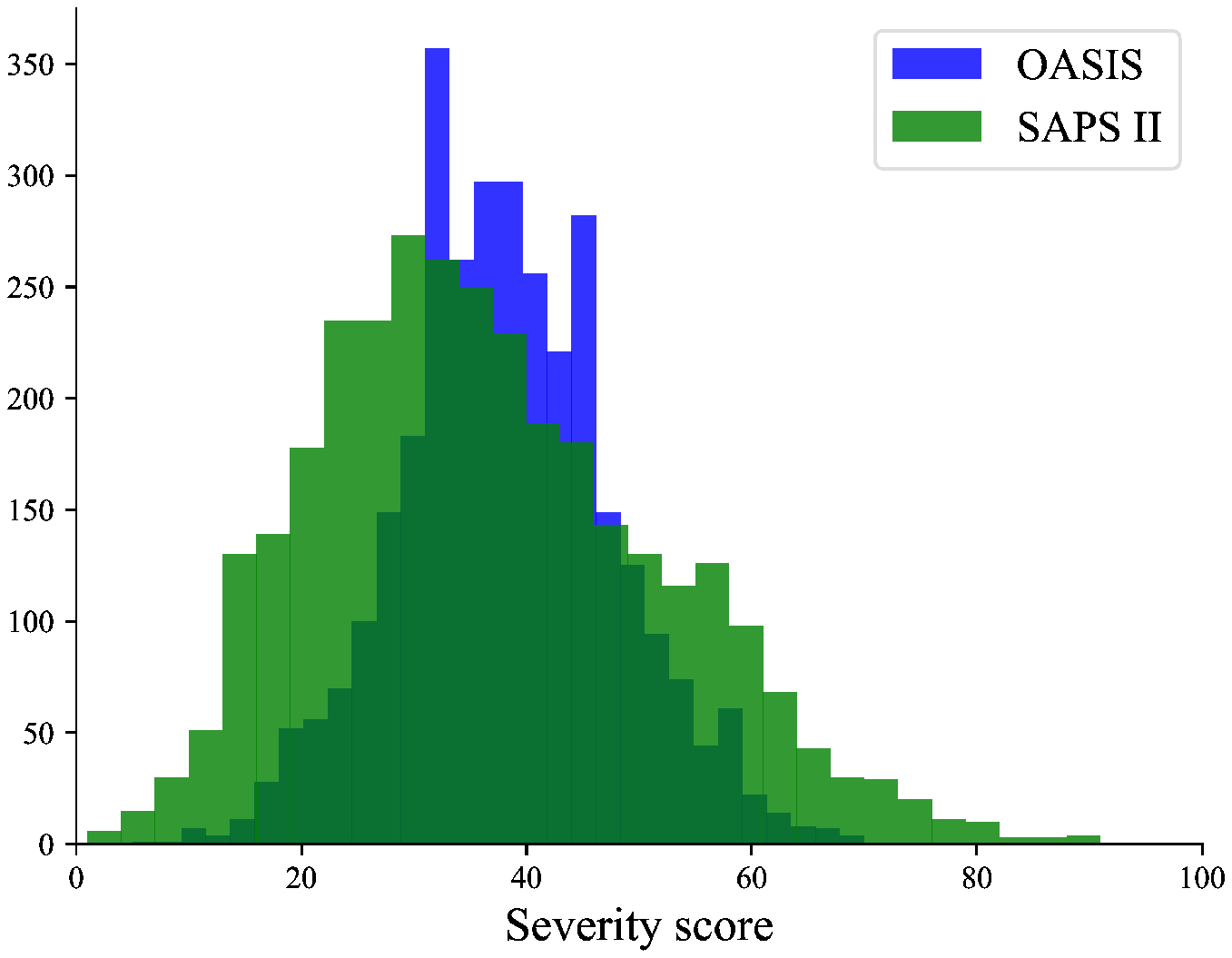}
        \caption{Test set score distribution.}
        \label{fig:test_score_dists}
    \end{subfigure}%
    \begin{subfigure}[b]{0.25\linewidth}   
        \centering 
        \includegraphics[width=\textwidth]{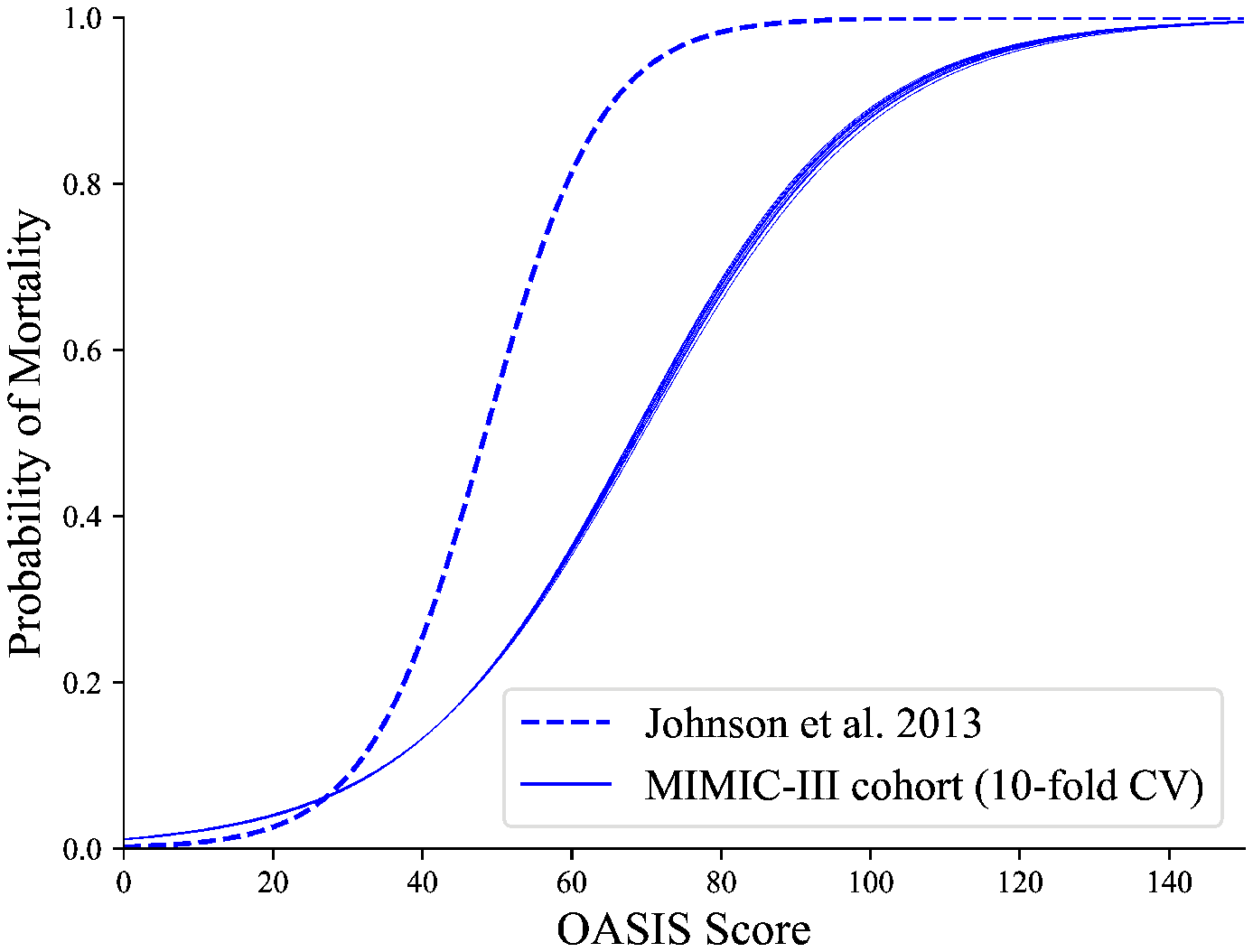}
        \caption{OASIS score calibration.}
        \label{fig:oasis_calibration}
    \end{subfigure}%
    \begin{subfigure}[b]{0.25\linewidth}   
        \centering 
        \includegraphics[width=\textwidth]{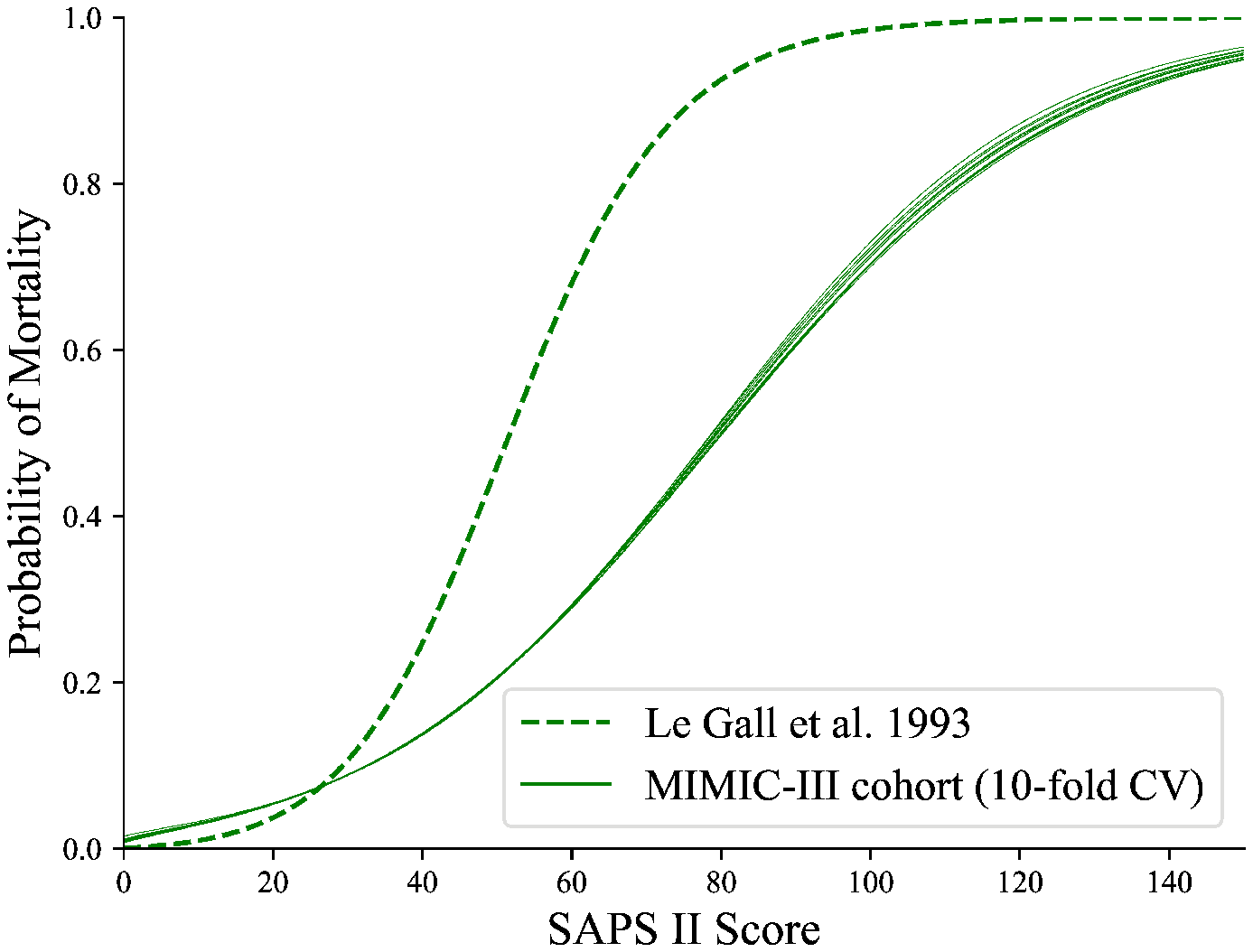}
        \caption{SAPS II score calibration.}
        \label{fig:saps2_calibration}
    \end{subfigure}
    \caption{Distribution and calibration of OASIS and SAPS II severity scores for the subset of the MIMIC III cohort who have sufficient data in the first 48 hours of their ICU stay showing that training and test score distributions are comparable (Panels a and b).}
    \label{fig:calibration}
\end{figure}

\begin{figure}
    \centering
    \begin{minipage}{0.5\textwidth}
        \includegraphics[width=\linewidth]{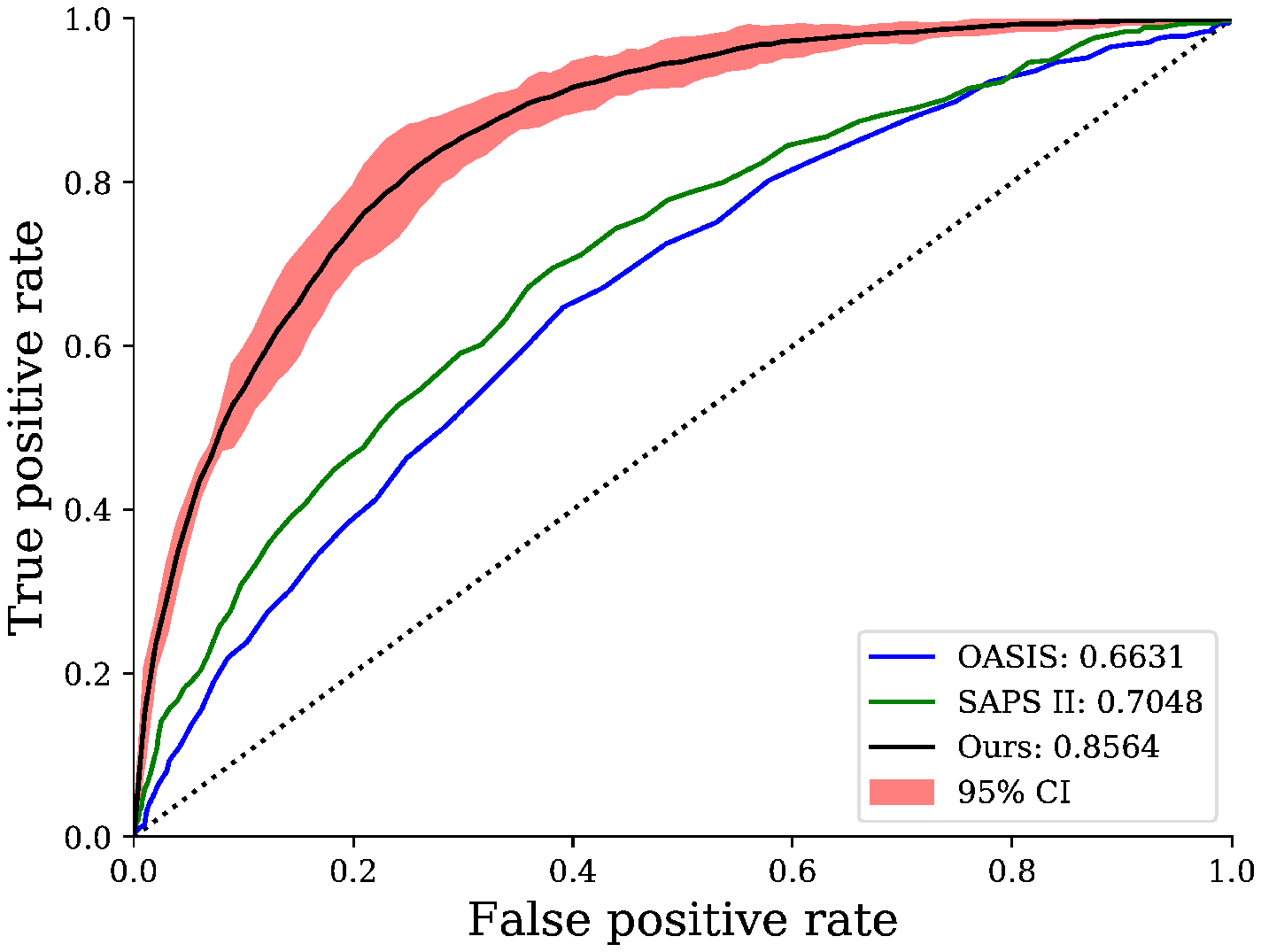}
    \end{minipage}%
    \begin{minipage}{0.5\textwidth}
        \includegraphics[width=\linewidth]{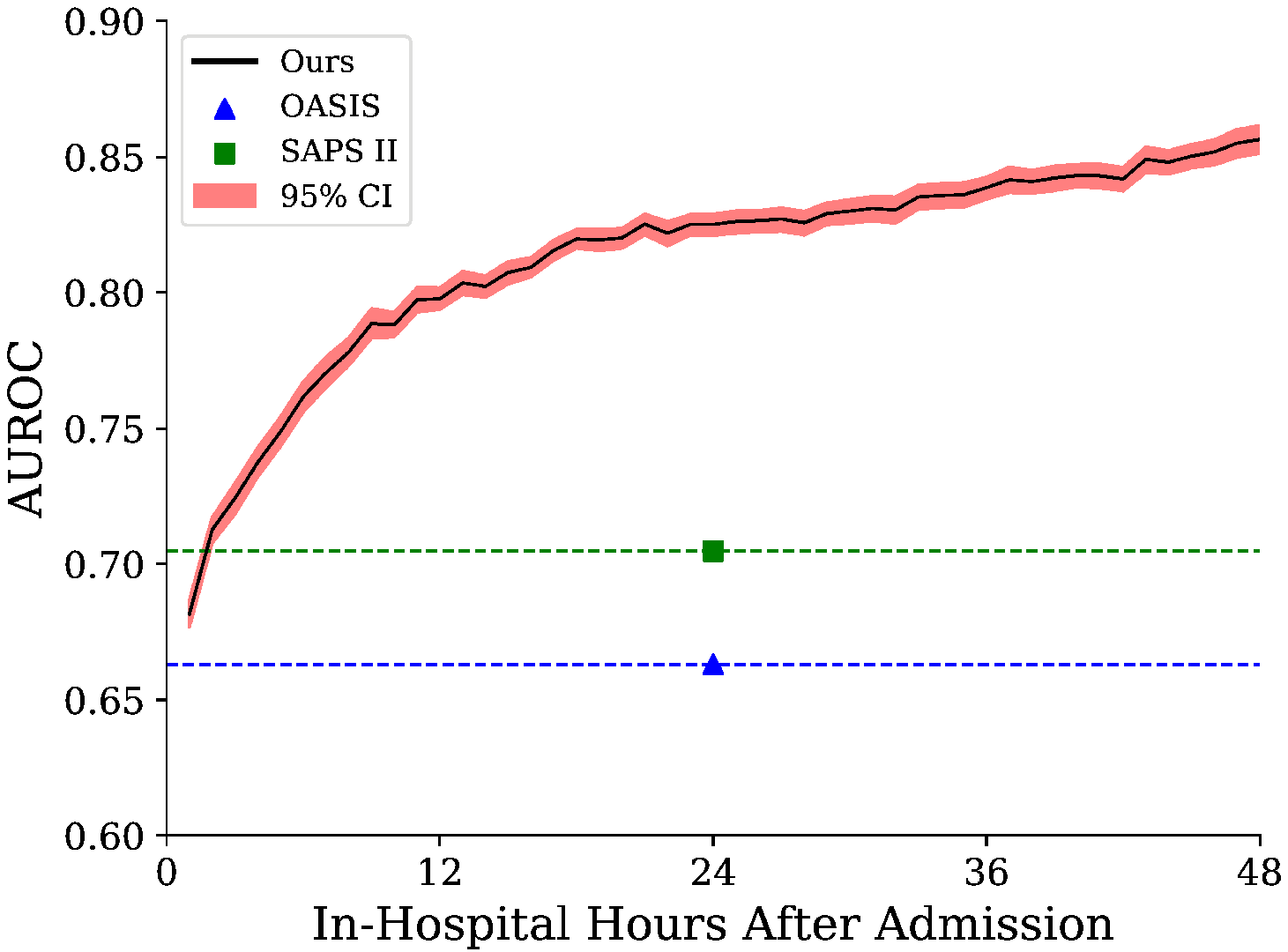}
    \end{minipage}
    \caption{Comparison of our model's dynamic prediction performance over 48 hours against the static prediction of OASIS and SAPS II (calculated at 24 hours after admission). Confidence intervals were bootstrapped from a 10-fold cross-validation. Left panel shows AUROC curves demonstrating significantly improved overall performance of our final model (at 48hours) compared to OASIS and SAPS II. The right panel shows the performance of the present model over time, demonstrating a performance exceeding that of the 24 hour OASIS and SAPS II models even within a few hours of admission and which continues to improve even up to 48 hours.}
    \label{fig:auroc_info}
\end{figure}

For comparison with our model, the OASIS and SAPS II severity scores were calculated on the training and test set. The distributions of both scores are depicted in Figure~\ref{fig:calibration}, demonstrating the similar spread of scores between the two sets. Function hyperparameters for converting from severity score to mortality probability were tuned on a 10-fold cross-validation of the training set using logistic regression and the Levenburg-Marquardt curve fitting algorithm \cite{more1978levenberg} for OASIS and SAPS II respectively. Figure~\ref{fig:calibration} shows the relatively shallow curves for our dataset in comparison to the respective original papers. As the difference between cross-validation folds when tuning was negligible, the resulting AUROC value differences were also negligible so we subsequently include only one fold of each severity score for clarity (see Figure~\ref{fig:auroc_info}).\\

For predicting inpatient mortality, the area under the receiver operating characteristic (AUROC) curve at 48 h after admission was 0$\cdot$8564 (95\% CI 0$\cdot$8512-0$\cdot$8614, bootstrapped from 10,000 samples of a 10-fold cross-validation). This was significantly stronger than traditional predictive models, with OASIS and SAPS II scores achieving 0$\cdot$6631 and 0$\cdot$7048 respectively. Figure~\ref{fig:auroc_info} illustrates the superior receiver operating characteristic of our model against both OASIS and SAPS II. After 48 hours of patient data has been accumulated, an AUROC value of 0$\cdot$86 means that  there is a 86\% chance our model will assign a higher probability to a randomly chosen patient destined to die rather than a randomly chosen patient destined to live. If clinical resource allocation were based on our model rather than SAPS II, approximately 25\% more patients would be correctly prioritised.\\

In Figure~\ref{fig:auroc_info}, we also illustrate the performance of our model through time - that is, the predictive strength of our model after each hour of a patient's stay. After the initial hour of patient stays, our model has an AUROC of 0$\cdot$68, better than the overall performance of the OASIS severity score. This is significant as OASIS relies on the entire first 24 hours of patient data and cannot make a prediction before or after this point. The same limitation applies to SAPS II, which is outperformed after 2 hours of patient data. Indeed, after only 12 hours, our model achieves an AUROC of over 0$\cdot$7977 (95\% CI 0$\cdot$7940-0$\cdot$8018) - performance which is arguably strong enough to assist the actions and prioritisation of clinicians. After the initial increase in model AUROC, the rate of improvement was found to become more incremental. Although our model continues to accumulate useful information for mortality prediction through time, new information at later stages in a patient's stay had less of an impact.\\

\noindent\textbf{DISCUSSION}\vspace{1mm}

\noindent We present an interpretable deep learning model using the entire multivariable patient time-series, regardless of variable type or frequency, and without the need for variable selection or cleaning. Our method capitalises on the flexible nature of word embeddings from NLP \cite{sutskever2014sequence,vaswani2017attention} and the success of RNNs in sequence analysis \cite{elman1990finding,rumelhart1988learning}, to greatly simplify the pre-processing required to use existing EHR structures for prediction. Our model is dynamic and able to track predicted survival probability for optimal prediction timing as well as providing an estimate of prediction confidence. As the vast majority of current techniques cannot make predictions at arbitrary times \cite{meiring2018optimal}, our time-sensitive model could help clinicians to assess overall patient trajectory, response to therapeutic interventions, guide optimal trials of intensive care, improve patient alerting and contribute to optimally informed shared decision-making conversations. This formulation leads to a significant improvement in the state-of-the-art for early patient outcome prediction when validating our method using the real-world ICU cohort in the MIMIC-III dataset \cite{johnson2016mimic}, and therefore may also find a role in benchmarking the quality of ICU care. Confident early prediction of low patient risk decreases the time needed for reassurance from traditional scoring systems and could be used as a guide for transfer from the ICU to a lower priority ward, potentially saving resources and staffing costs\\

Unlike \cite{rajkomar2018scalable}, our model represents variables in a single large embedding space and is, to our knowledge, the first time all clinical variables from an EHR database have been represented in the same latent space. Therefore, our model is the first attempt to teach an ``AI Clinician'' \cite{komorowski2018artificial} to relate all EHR variables on the same basis -- analogous to asking a healthcare provider to relate everything from observations to heart rate in an unbiased way. Complex machine learning techniques (and deep learning in particular) can be opaque, and this has been a criticism in the medical domain where decision-making must be transparent to be acceptable to both patients and clinicians. However, our model design inherently provides a degree of insight by ranking the relevance of clinical variables. As such, our results aid clinicians in focusing their attention across all clinical variables, help treatment decision making and demonstrate surprisingly important factors in patient outcome prediction.\\

We created a processing and embedding technique that assimilated all discrete and continuous events in each patient's EHR because we were interested in generalising the embedding mechanism demonstrated in \cite{rajkomar2018scalable} to arbitrary EHR formats. Preservation of so many variable types allows our model to learn from a far broader range of ICU data than previous models. For instance, the model now assesses nursing notes at the same time as checking the most recent laboratory values; an experience much closer to that of a clinician. Our model also has the capacity to relate unusual or infrequently sampled events across time - insight that clinicians in the highly demanding setting of the ICU may struggle to appreciate. As the spread of EHR systems proliferates, larger datasets employing models with this type of flexible embedding could also lead to clinical and physiological insights beyond those of the current medical corpus. This superhuman comprehension of diverse data has already been demonstrated for image classification \cite{he2016deep,szegedy2016rethinking} and the data dense setting provided by modern EHR systems is likely to contain similar complex insights.\\

\begin{figure}
    \centering
    \begin{minipage}{0.5\textwidth}
        \includegraphics[width=\linewidth]{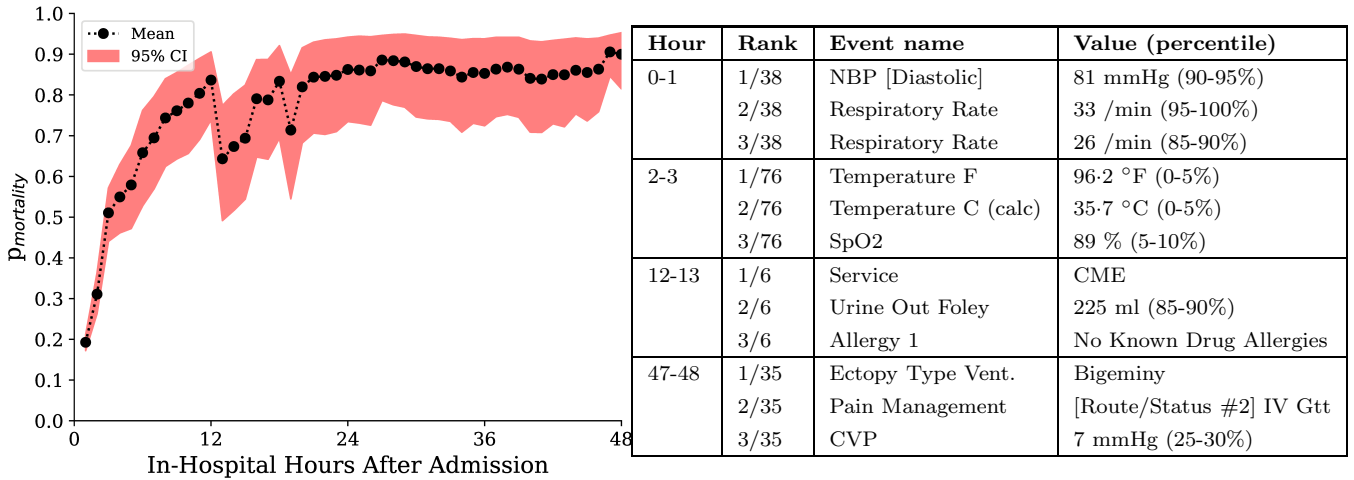}
    \end{minipage}\hfill%
    \begin{minipage}{0.5\textwidth}
        \renewcommand*{\arraystretch}{1.3}
        \footnotesize
        \begin{tabular}{|l|l|l|l|}
            \hline
            \textbf{Hour} & \textbf{Rank} & \textbf{Event name} & \textbf{Value (percentile)} \\ \hline
            0-1   & 1/38 & NBP [Diastolic]        & 81 mmHg (90-95\%) \\
                  & 2/38 & Respiratory Rate       & 33 /min (95-100\%) \\
                  & 3/38 & Respiratory Rate       & 26 /min (85-90\%) \\ \hline
            2-3   & 1/76 & Temperature F          & 96$\cdot$2 $^{\circ}$F (0-5\%) \\
                  & 2/76 & Temperature C (calc)   & 35$\cdot$7 $^{\circ}$C (0-5\%) \\
                  & 3/76 & SpO2                   & 89 \% (5-10\%) \\ \hline
            12-13 & 1/6  & Service                & CME \\
                  & 2/6  & Urine Out Foley        & 225 ml (85-90\%) \\
                  & 3/6  & Allergy 1              & No Known Drug Allergies \\ \hline
            47-48 & 1/35 & Ectopy Type Vent.      & Bigeminy \\
                  & 2/35 & Pain Management        & [Route/Status \#2] IV Gtt \\
                  & 3/35 & CVP                    & 7 mmHg (25-30\%) \\ \hline
        \end{tabular}
    \end{minipage}
    \caption{Dynamic probability of mortality after ICU admission for a patient who subsequently died during their stay in hospital. Event rank within the each hour, event name and event value (percentile) are shown for the first and last hour, as well as hours where there is a significant change in likelihood.}
    \label{case_study_1}
\end{figure}

The combination of dynamic, individual mortality predictions with calibrated uncertainty provides a summary of the patient state which may be useful in automated alerting and clinician prioritisation. The time-sensitivity means that it may be used to track patient trajectory as well as evaluate response to therapy. The availability of ranked salient features is a step towards providing the interpretability needed to make such systems acceptable both to patients or patient advocates and clinicians in guiding individualised care. For example, Figure~\ref{case_study_1} shows hourly predictions for a patient, a male of over 90 years of age on his 2nd stay in the ICU after being discharged 2 months before, who died of congestive heart failure 65 hours after admission. Our model made an initial prediction after one hour of patient data, predicting that the patient had a 19$\cdot$2\% chance of mortality. This is already a significant deviation from the population average mortality rate of 13$\cdot$2\% and was based on the importance our model places on high blood pressure and high respiratory rate. Our model later indicates the patient deteriorating to over 50\% risk of death at hour 3, due to worryingly low temperature and SpO$_2$. Assessment of our model's predictions at this point in time would have already shown a very concerning trend in patient risk, potentially reinforcing clinician suspicions at the time. Subsequently, after 12 hours, over 2 days before death occurred, our model gave the patient an 83$\cdot$6\% chance of death. By the time a traditional severity score such as OASIS or SAPS II could have been used to calculate patient mortality risk, our model had already indicated this patient was at incredibly high risk for several hours. At the end of our prediction window the likelihood of survival was less than 10\%.\\

Figure~\ref{case_study_1} also highlights the limitations of our model. In the 13$^{\textrm{th}}$ hour of the patient's stay, when there are only 6 readings taken, it is unclear why the model prioritises \emph{Service CME} and \emph{Allergy 1 No Known drug allergies}. As the model is broad enough to process all clinical variables, it will also rank many superfluous records that could affect prediction. Another limitation of the study is the inherent bias introduced when obtaining quantiles from specific hospital cohorts. This renders our model subject to both hospital-based and demographic-based bias. Moreover, the principle concern with maximally flexible models that can intake all forms of input, including clinician notes, is that models will learn to extract trivial identifiers of impending outcomes. This could include discharge notes being indicative of survival, or clinician comments on patient outcomes being directly used for prediction.  In our experiments we found events such as recording the code status instructions for cpr, assigning consent to a next of kin, and a visitation by the priest to all be highly indicative of mortality. By employing clearly interpretable models, such as our ranking system, a clinician could ignore such tautologies and consider whether they have fully considered the implications of our model's next most important events.\\

We believe our model to be the first that performs no variable selection or pre-processing as well as allowing for erroneous values. Recent work with EHR data \cite{rajkomar2018scalable,tabak2013using} has focused on scalability and streamlining the transition between widespread data formats and model inputs. Our model goes one step further by allowing for all types of readings to be used as inputs and assessed for any correlation with patient outcomes. Furthermore, when employing our pipeline, there is no necessary censoring of patients during secondary analysis of EHR data. We present a deep learning model which can generate time-sensitive mortality probability estimates as a summary measure of patient state with calibrated confidence estimates for an individual patient at any arbitrary time. The model is able to assimilate \emph{all} the data available without the need for cleaning. Unlike previous ICU prediction models, we treat all variables on the same basis, without the need for feature engineering, using a single large embedding space. Even without pre-processing, we achieve state-of-the-art performance across the cohort. Our approach is a natural way to handle the complex structure of ICU data and provides a summary of patient state over time, as well as making the salient features available which may be useful in guiding clinicians.\\



\noindent\textbf{Contributors}\vspace{1mm}\\
JD, AE and PL conceived the idea. JD undertook the modelling and analysis. All authors contributed to the writing of the manuscript.\\

\noindent\textbf{Declaration of interests}\vspace{1mm}\\
All authors declare no competing interests.\\

\noindent\textbf{Data sharing}\vspace{1mm}\\
The MIMIC-III database is part of restricted-access clinical data maintained by PhysioNet (MIMIC-II, MIMIC-III, eICU Collaborative Research Database) and is available, subject to a formal research request, from the MIT Laboratory for Computational Physiology and their collaborating research groups \href{https://mimic.physionet.org/gettingstarted/access/}{https://mimic.physionet.org/gettingstarted/access/}. The code used for these experiments is available at \href{https://github.com/jacobdeasy/flexible-ehr}{https://github.com/jacobdeasy/flexible-ehr}.\\

\noindent\textbf{Acknowledgements}\vspace{1mm}\\
This work was supported by the UK Medical Research Council and Raymond and Beverley Sackler Foundation.\\

\noindent\textbf{Role of the funding source}\\
The funders of the study had no role in study design, data collection, data analysis, data interpretation, or writing of the report. The corresponding author had full access to all the data in the study and had final responsibility for the decision to submit for publication.

\bibliographystyle{vancouver}
\bibliography{refs}

\begin{thebibliography}{10}

\bibitem{desautels2017prediction}
Desautels T, Das R, Calvert J, Trivedi M, Summers C, Wales DJ, et~al.
\newblock Prediction of early unplanned intensive care unit readmission in a UK
  tertiary care hospital: a cross-sectional machine learning approach.
\newblock BMJ open. 2017;7(9):e017199.

\bibitem{graham2010monitor}
Graham KC, Cvach M.
\newblock Monitor alarm fatigue: standardizing use of physiological monitors
  and decreasing nuisance alarms.
\newblock American Journal of Critical Care. 2010;19(1):28--34.

\bibitem{drew2014insights}
Drew BJ, Harris P, Z{\`e}gre-Hemsey JK, Mammone T, Schindler D, Salas-Boni R,
  et~al.
\newblock Insights into the problem of alarm fatigue with physiologic monitor
  devices: a comprehensive observational study of consecutive intensive care
  unit patients.
\newblock PloS one. 2014;9(10):e110274.

\bibitem{goldstein2017opportunities}
Goldstein BA, Navar AM, Pencina MJ, Ioannidis J.
\newblock Opportunities and challenges in developing risk prediction models
  with electronic health records data: a systematic review.
\newblock Journal of the American Medical Informatics Association.
  2017;24(1):198--208.

\bibitem{johnson2016machine}
Johnson AE, Ghassemi MM, Nemati S, Niehaus KE, Clifton DA, Clifford GD.
\newblock Machine learning and decision support in critical care.
\newblock Proceedings of the IEEE Institute of Electrical and Electronics
  Engineers. 2016;104(2):444.

\bibitem{che2018recurrent}
Che Z, Purushotham S, Cho K, Sontag D, Liu Y.
\newblock Recurrent neural networks for multivariate time series with missing
  values.
\newblock Scientific reports. 2018;8(1):6085.

\bibitem{meiring2018optimal}
Meiring C, Dixit A, Harris S, MacCallum NS, Brealey DA, Watkinson PJ, et~al.
\newblock Optimal intensive care outcome prediction over time using machine
  learning.
\newblock PloS one. 2018;13(11):e0206862.

\bibitem{rajkomar2018scalable}
Rajkomar A, Oren E, Chen K, Dai AM, Hajaj N, Hardt M, et~al.
\newblock Scalable and accurate deep learning with electronic health records.
\newblock NPJ Digital Medicine. 2018;1(1):18.

\bibitem{johnson2016mimic}
Johnson AE, Pollard TJ, Shen L, Li-wei HL, Feng M, Ghassemi M, et~al.
\newblock MIMIC-III, a freely accessible critical care database.
\newblock Scientific data. 2016;3:160035.

\bibitem{harutyunyan2017multitask}
Harutyunyan H, Khachatrian H, Kale DC, Galstyan A.
\newblock Multitask learning and benchmarking with clinical time series data.
\newblock arXiv preprint arXiv:170307771. 2017;.

\bibitem{malone2018learning}
Malone B, Garcia-Duran A, Niepert M.
\newblock Learning Representations of Missing Data for Predicting Patient
  Outcomes.
\newblock arXiv preprint arXiv:181104752. 2018;.

\bibitem{srivastava2014dropout}
Srivastava N, Hinton G, Krizhevsky A, Sutskever I, Salakhutdinov R.
\newblock Dropout: a simple way to prevent neural networks from overfitting.
\newblock The Journal of Machine Learning Research. 2014;15(1):1929--1958.

\bibitem{gal2016theoretically}
Gal Y, Ghahramani Z.
\newblock A theoretically grounded application of dropout in recurrent neural
  networks.
\newblock In: Advances in neural information processing systems; 2016. p.
  1019--1027.

\bibitem{kingma2014adam}
Kingma DP, Ba J.
\newblock Adam: A method for stochastic optimization.
\newblock arXiv preprint arXiv:14126980. 2014;.

\bibitem{efron1994introduction}
Efron B, Tibshirani RJ.
\newblock An introduction to the bootstrap.
\newblock CRC press; 1994.

\bibitem{more1978levenberg}
Mor{\'e} JJ.
\newblock The Levenberg-Marquardt algorithm: implementation and theory.
\newblock In: Numerical analysis. Springer; 1978. p. 105--116.

\bibitem{sutskever2014sequence}
Sutskever I, Vinyals O, Le QV.
\newblock Sequence to sequence learning with neural networks.
\newblock In: Advances in neural information processing systems; 2014. p.
  3104--3112.

\bibitem{vaswani2017attention}
Vaswani A, Shazeer N, Parmar N, Uszkoreit J, Jones L, Gomez AN, et~al.
\newblock Attention is all you need.
\newblock In: Advances in neural information processing systems; 2017. p.
  5998--6008.

\bibitem{elman1990finding}
Elman JL.
\newblock Finding structure in time.
\newblock Cognitive science. 1990;14(2):179--211.

\bibitem{rumelhart1988learning}
Rumelhart DE, Hinton GE, Williams RJ, et~al.
\newblock Learning representations by back-propagating errors.
\newblock Cognitive modeling. 1988;5(3):1.

\bibitem{komorowski2018artificial}
Komorowski M, Celi LA, Badawi O, Gordon AC, Faisal AA.
\newblock The Artificial Intelligence Clinician learns optimal treatment
  strategies for sepsis in intensive care.
\newblock Nature Medicine. 2018;24(11):1716.

\bibitem{he2016deep}
He K, Zhang X, Ren S, Sun J.
\newblock Deep residual learning for image recognition.
\newblock In: Proceedings of the IEEE conference on computer vision and pattern
  recognition; 2016. p. 770--778.

\bibitem{szegedy2016rethinking}
Szegedy C, Vanhoucke V, Ioffe S, Shlens J, Wojna Z.
\newblock Rethinking the inception architecture for computer vision.
\newblock In: Proceedings of the IEEE conference on computer vision and pattern
  recognition; 2016. p. 2818--2826.

\bibitem{tabak2013using}
Tabak YP, Sun X, Nunez CM, Johannes RS.
\newblock Using electronic health record data to develop inpatient mortality
  predictive model: Acute Laboratory Risk of Mortality Score (ALaRMS).
\newblock Journal of the American Medical Informatics Association.
  2013;21(3):455--463.

\end{thebibliography}

\end{document}